\documentclass[conference]{IEEEtran}
\IEEEoverridecommandlockouts

\usepackage{cite}
\usepackage{amsmath,amssymb,amsfonts}
\usepackage{algorithm}
\usepackage{algpseudocode}
\usepackage{pgfplots}
\usepackage{textcomp}
\usepackage{xcolor}
\usepackage{tikz}
\usepackage{graphicx}
\usepackage{url}
\usepackage{expl3}
\usepackage{glossaries}
\usepackage{subcaption}
\usepackage{float}
\usepackage{bm}

\newglossary[mlg]{math}{mls}{mln}{Mathematical Terms}

\newcommand{\loss}{L}

\newcommand{\sampleidx}{r} 
\newcommand{\truelabel}{y}
\newcommand{\hypothesis}{h} 
\newcommand{\vx}[0]{{\bf x}}
\newcommand{\featurevec}{\vx}

\newcommand{\hypospace}{\mathcal{H}}
\newcommand{\samplesize}{m}
\newcommand{\dataset}{\mathcal{D}}

\def\BibTeX{{\rm B\kern-.05em{\sc i\kern-.025em b}\kern-.08em
    T\kern-.1667em\lower.7ex\hbox{E}\kern-.125emX}}
\begin{document}

\title{Interpretable Multiple Myeloma Prognosis with Observational Medical Outcomes Partnership Data
\thanks{This work was partially supported by the Research Council of Finland under grants 349966, 363624, the 
Jane and Aatos Erkko Foundation under grant A835.}
}

\author{\IEEEauthorblockN{Salma Rachidi}
\IEEEauthorblockA{
\textit{Aalto University} \\
\textit{Dept. of Computer Science} \\
Espoo, Finland\\
0009-0002-5132-485X}
\and
\IEEEauthorblockN{Aso Bozorgpanah}
\IEEEauthorblockA{\textit{Aalto University} \\
	\textit{Dept. of Computer Science} \\
Espoo, Finland\\
0009-0007-1502-0927}

\and
\IEEEauthorblockN{Eric Fey}
\IEEEauthorblockA{\textit{Helsinki University Hospital} \\
	Helsinki, Finland\\
	0009-0005-7847-7479
	}

\and
\IEEEauthorblockN{Alexander Jung}
\IEEEauthorblockA{\textit{Aalto University} \\
	\textit{Dept. of Computer Science} \\
	Espoo, Finland\\
	0000-0001-7538-0990
	}}

\maketitle

\begin{abstract}
Machine learning (ML) promises better clinical decision-making, yet 
opaque model behavior limits the adoption in healthcare. 
We propose two novel regularization techniques for ensuring the 
interpretability of ML models trained on real-world data. 
In particular, we consider the prediction of five-year survival for multiple myeloma patients using clinical data from Helsinki University Hospital.
To ensure the interpretability of the trained models, we use two 
alternative constructions for a penalty term used for regularization. 
The first one penalizes deviations from the predictions obtained 
from an interpretable logistic regression method with two manually 
chosen features. The second construction requires consistency of model 
predictions with the revised international staging system (R-ISS).
We verify the usefulness of the proposed regularization techniques in 
numerical experiments using data from 812 patients. They achieve an accuracy up to 0.721 on a test set and SHAP values show that the 
models rely on the selected important features.
\end{abstract}

\begin{IEEEkeywords}
interpretable ML, survival prediction, clinical ML, mulitple myeloma.
\end{IEEEkeywords}

\newcommand{\meanpm}[2]{\num{#1}\,$\pm$\,\num{#2}}
\def\vx{\mathbf{x}}
\def\L{{\cal L}}
\newcommand{\simpleprox}{g^{*}}
\newcommand{\auxmodel}{g^{*}}

\section{Introduction}

Machine learning (ML) methods have increasing potential to support clinical decision making \cite{ml_potential}. In healthcare applications,
predictive accuracy alone is insufficient: ML systems must also be
interpretable to enable trustworthy decision making
\cite{goals}. While expressive ML models can
capture complex dependencies in clinical data, they often operate as black
boxes, making their predictions difficult to justify and validate
clinically \cite{Amann2020Explainability}.

Most existing approaches address interpretability using post-hoc
explanation methods, such as feature attribution 
\cite{Ribeiro2016LIME,Lundberg2017SHAP}. Although useful for model analysis,
these techniques do not influence the training process \cite{molnar} and provide no
guarantees that the learned decision function is clinically
meaningful or stable. 

Multiple myeloma (MM) is a malignancy of plasma cells that predominantly affects older
patients, and survival outcomes depend on a
heterogeneous set of clinical variables, including laboratory
measurements, staging scores, cytogenetic markers, and treatment-related
factors \cite{abduh}. 

In this work, we adopt a training-time approach to interpretability by
explicitly incorporating clinically motivated constraints into the
learning objective of models predicting prognosis of MM patients. This perspective builds on our recent work on
\emph{explainable empirical risk minimization} (EERM), which formalizes
interpretability as a regularization term in the training objective
\cite{eerm}. We construct regularizers that penalize
deviations from desired interpretable behavior, thereby shaping the
optimization landscape so that the resulting models are interpretable
by design.

{\bf Contributions.} This paper makes two main contributions.
\begin{itemize}
	\item We propose two training-time regularizers for interpretability.
	The first aligns a flexible neural network with an auxiliary,
	intrinsically interpretable model trained on clinically selected
	features. The second enforces stage-consistent predictions based on the
	Revised International Staging System (R-ISS) \cite{riss}, embedding established
	clinical risk stratification directly into the learning objective.
	\item We apply the proposed regularization framework to real-world MM
	data extracted from an Observational Medical Outcomes Partnership (OMOP) Common Data Model (CDM) database at Helsinki University Hospital (HUS) \cite{bergius}, demonstrating that interpretable-by-design	models can achieve competitive predictive performance while exhibiting
	clinically coherent behavior.
\end{itemize}

{\bf Related work.}
Several studies have applied machine learning to survival prediction in MM, 
using different ML models such as logistic regression, random forests, and 
gradient-boosted trees \cite{georgoula,belmonte,chen}. Interpretability in 
these works is typically addressed using post-hoc explanation methods such 
as SHAP or permutation-based feature importance. In contrast, our approach 
incorporates interpretability directly into the training objective via 
explicit regularization, yielding models that are interpretable by design.

\section{Problem Setting} 
\label{problem_setting}

We consider the problem of predicting five-year survival for patients diagnosed with
MM using routinely collected clinical data. Each patient is represented by a feature 
vector comprising demographic information and laboratory measurements, and
the prediction target is a binary label indicating whether the patient died within five
years from diagnosis. 

{\bf Simple models as auxiliary structure.}
Simple and intrinsically interpretable models, such as logistic regression models trained
on a small number of carefully selected features, are often used in clinical practice
as baseline risk predictors \cite{wolfrath}. While such models may be limited 
in expressive power, they provide transparent and clinically plausible decision rules. 
In this work, we treat the predictions of such interpretable models as auxiliary 
signals that encode domain knowledge. These signals are not used as final predictors, 
but rather to guide the training of more expressive models through explicit regularization.

{\bf Clinical staging as domain structure.} Prognosis in MM is commonly stratified 
using the Revised International Staging System (R-ISS), which combines beta-2 microglobulin 
(B2M), albumin, lactate dehydrogenase (LDH), and cytogenetic abnormalities to define three 
clinically meaningful risk stages \cite{riss}. Patients assigned to the same stage are 
expected to exhibit broadly similar survival behavior. In this work, we exploit this 
clinically established structure as a source of domain knowledge for constructing 
interpretability-driven regularization terms that encourage clinically coherent model 
predictions.

\section{Dataset} \label{dataset}

{\bf Data source.} The data are extracted from a clinical database at 
Helsinki University Hospital (HUS) structured according to the OMOP CDM. This CDM was 
developed by the Observational Health Data Sciences and Informatics (OHDSI) 
community to provide a standardized representation of clinical data and 
facilitate federated analysis across institutions \cite{bergius}.

{\bf Cohort and prediction target.} Each data point corresponds to a patient diagnosed with MM and treated at HUS. The prediction target (or label) is five-year survival from the date of diagnosis, represented as a binary label indicating whether the patient died within five years. Only patients with known 5-year survival (i.e. at least 5-year follow-up for censored patients) and available values for beta-2 microglobulin (B2M), albumin, and lactate dehydrogenase (LDH), the core components of the R-ISS, are included. 

{\bf Features.}
The feature vector includes
(i) age at diagnosis and
(ii) biomarker values measured within a time window of $-30$ to $+30$ days from diagnosis,
where the most recent available measurement within this window is used. The selected
features reflect established prognostic factors in multiple myeloma as well as prior clinical and
ML studies \cite{riss,georgoula, evolution, index, alp}. All patients have known age at diagnosis. Missing values in other features are imputed using the mean of the corresponding feature computed on the training data.

A summary of the resulting dataset is provided in Table~\ref{table:myeloma_HUS}.

\begin{table}[h]
	\centering
	\caption{Summary of the dataset}
	\label{table:myeloma_HUS}
	\begin{tabular}{|p{1.5cm}|p{5.7cm}|}
		\hline
		\textbf{Aspect} & \textbf{Description} \\
		\hline
		Data Points & 812 myeloma patients treated at HUS who have known 5-year 
		survival status and available values for albumin, LDH and B2M \\
		\hline
		Features  & Age at diagnosis, 16 blood measurements 
		(albumin, LDH, B2M, hemoglobin, platelets, creatinine, C reactive protein, 
		alkaline phosphatase, free lambda light chains, free kappa light chains, 
		leukocytes, protein, ionized calcium, immunoglobulin A, 
		immunoglobulin G and immunoglobulin M) and percentage of 
		plasma cells in bone marrow\\
		\hline
		Label & Binary : 1 if the patient died within 5 years from diagnosis, 
		0 if the patient survived at least 5 years\\
		\hline
		Data \newline distribution & 417 data points with label 1 and 395 with label 0 \newline stage 1: 18/111 (16.2 \% ) with label 1
		\newline stage 2: 325/595 (54.6 \%) with label 1
		\newline stage 3: 74/106 (69.8 \%) with label 1 \\
		\hline
	\end{tabular}
\end{table}

\section{Methodology} \label{method}

We study two constructions of regularizers that enforce interpretability of ML models directly during training. The proposed regularizers are model agnostic 
and can be combined with any hypothesis class trained via empirical risk 
minimization. Moreover, they are independent of the specific choice of loss function and
can be used with a wide range of training objectives. In this work, we instantiate the
approach using artificial neural networks (ANNs) and the logistic loss to demonstrate
compatibility with expressive, high-capacity models in a standard classification setting.

In both constructions, we assume that a predictor for MM survival is learned. 
The first construction aligns the predictor with an auxiliary, intrinsically 
interpretable model for MM survival. The second construction enforces consistency 
of predictions within R-ISS stages.

Given a labeled training set
$\{(\featurevec^{(\sampleidx)}, \truelabel^{(\sampleidx)})\}_{\sampleidx=1}^{\samplesize}$,
we learn a predictor $\hat{\hypothesis} \in \hypospace$ by solving
\begin{equation}
\label{equ_def_rerm}
\hat{\hypothesis} \in \operatorname{arg\,min}_{\hypothesis \in \hypospace}
\bigg[
\frac{1}{\samplesize} \sum_{\sampleidx=1}^{\samplesize}
\loss\!\left(\hypothesis(\featurevec^{(\sampleidx)}), \truelabel^{(\sampleidx)}\right)
+ \alpha \, \mathcal{R}(\hypothesis)
\bigg].
\end{equation}
The main contribution of this work are two interpretability-driven 
constructions of a regularizer $\mathcal{R}(\hypothesis)$.

Note that our approach is agnostic to the choice of loss function $\loss$ and 
model $\hypospace$ used in \eqref{equ_def_rerm}. In our experiments we train an 
ANN using logistic loss \cite[Ch. 3]{JungMLBasics}. The regularization parameter $\alpha \geq 0$ 
in \eqref{equ_def_rerm} controls the strength of the interpretability constraint: 
$\alpha = 0$ means no interpretability requirement, while larger values of 
$\alpha$ increasingly enforce clinical interpretability.

\subsection{Auxiliary Alignment Regularizer} \label{aux_model}

The first regularizer $\mathcal{R}^{(\rm AA)}(\hypothesis)$ leverages a simple, intrinsically 
interpretable auxiliary model $g$. This auxiliary model is pre-trained and delivers 
a soft label $g(\featurevec) \in (0,1)$ for a patient with features $\featurevec$. 
We can interpret $g(\featurevec)$ as a risk score assigned by the auxiliary model 
to a patient with features $\featurevec$. 

The regularizer $\mathcal{R}^{(\rm AA)}(\hypothesis)$ is then obtained by
measuring the deviation between the predictions of the learned model 
$\hypothesis$ and the auxiliary model $g$ on a given dataset $\mathcal{D}$. 
In particular, we use the Kullback--Leibler (KL) divergence between the 
Bernoulli distributions induced by $\hypothesis(\featurevec)$ and $g(\featurevec)$:
\begin{equation}
	\label{eq:aux_reg}
\begin{split}
\mathcal{R}^{(\rm AA)}(\hypothesis) &=
\frac{1}{|\mathcal{D}|} \sum_{\featurevec \in \mathcal{D}}
\Bigg[
\widehat{g}(\featurevec) \log
\!\left(
\frac{\widehat{g}(\featurevec)}{\hypothesis(\featurevec)}
\right) \\
&\quad + \big(1 - \widehat{g}(\featurevec)\big)
\log
\!\left(
\frac{1 - \widehat{g}(\featurevec)}
     {1 - \hypothesis(\featurevec)}
\right)
\Bigg].
\end{split}
\end{equation}
We stress that this construction can be applied to any model 
$\hypothesis \in \hypospace$ and auxiliary model $\widehat{g}$ that deliver soft 
label values in $(0,1)$. For example, $g$ or $\hypothesis$ could 
be obtained from decision trees or random forests. 

In our experiments (see Section \ref{results}), the auxiliary model 
$\widehat{g}$ is obtained from logistic regression trained on a small 
number features. To select these features, we perform an exhaustive 
search over all possible pairs of candidate features. The feature pair 
yielding the lowest validation loss is selected. 

\subsection{Stage-Consistency Regularizer} \label{stage}

The second regularizer, denoted $\mathcal{R}^{(\rm SC)}(\hypothesis)$, 
exploits clinically established R-ISS staging information. Each patient is 
assigned to a stage based on routinely available biomarkers using the
procedure summarized in Algorithm~\ref{alg_r_iss}, adapted to the absence 
of cytogenetic information in the available data \cite{riss,rachidi}. 
The thresholds for normal and elevated LDH values follow reference ranges 
used at HUS \cite{hus_ldh}.

Let $\dataset^{(s)}$ denote the subset of patients in the training set with 
stage $s$. The stage-consistency regularizer penalizes deviations of individual 
predictions from the mean outcome within each stage:
\begin{align}
	\label{equ:stage_reg}
\mathcal{R}^{(\rm SC)}(\hypothesis)
&= \sum_{s=1}^{S}
\frac{1}{|\dataset^{(s)}|}
\sum_{\featurevec \in \dataset^{(s)}}
\big( \hypothesis(\featurevec) - \mu^{(s)} \big)^2, \nonumber \\
\mu^{(s)} &=
\frac{1}{|\dataset^{(s)}|}
\sum_{\sampleidx:\featurevec^{(\sampleidx)} \in \dataset^{(s)}}
\truelabel^{(\sampleidx)} .
\end{align}
Like $\mathcal{R}^{(\rm AA)}(\hypothesis)$, also $\mathcal{R}^{(\rm SC)}(\hypothesis)$ 
can be combined with any model $\hypothesis \in \hypospace$.

\begin{algorithm}[h]
\caption{Staging with R-ISS}
\label{alg_r_iss}
\begin{algorithmic}[1]
  \State \textbf{Input:} B2M, LDH, albumin, and age of a patient.
  \If{age $< 70$}
      \State $high\_ldh \gets (LDH > 235)$
  \Else
      \State $high\_ldh \gets (LDH > 255)$
  \EndIf
  \If{B2M $< 3.5$ and albumin $\geq 35$ and not $high\_ldh$}
      \State $stage \gets 1$
  \ElsIf{B2M $\geq 5.5$ and $high\_ldh$}
      \State $stage \gets 3$
  \Else
      \State $stage \gets 2$
  \EndIf
  \State \Return $stage$
\end{algorithmic}
\end{algorithm}

\section{Numerical Experiments} \label{results}

Our numerical experiments evaluate whether the regularizers introduced in
Section~\ref{method} can steer model training in a clinically meaningful way 
without degrading predictive performance. We study this trade-off 
in the MM five-year survival prediction task described in
Section~\ref{problem_setting}.

We first establish baseline performance, and then analyze how varying
the regularization strength $\alpha$ in \eqref{equ_def_rerm} affects
predictive accuracy, AUC, loss decomposition, and feature attribution
for both regularization strategies.

\subsection{Data splits and evaluation protocol}

The data points are randomly partitioned into three disjoint subsets:

\begin{enumerate}
	\item $\mathcal{D}^{(\rm aux)}$, consisting of 122 patient records,
	      used exclusively to estimate the parameters of the auxiliary
	      logistic regression model $\widehat{g}$.
	\item $\mathcal{D}^{(\rm kf)}$, consisting of 568 patient records,
	      used for $k$-fold cross-validation, hyperparameter tuning, and
	      training of the final ANN models $\hypospace$ in \eqref{equ_def_rerm}. 
		  This dataset is also used to select the two features used for the auxiliary model $\widehat{g}$ with respect to its loss values using exhaustive search. 
	\item $\mathcal{D}^{(\rm test)}$, consisting of 122 patient records,
	      solely reserved for final performance and
	      interpretability evaluation.
\end{enumerate}

Hyperparameters of the ANN architecture and optimizer are selected by
minimizing the mean validation loss over the $k$ folds using
$\mathcal{D}^{(\rm kf)}$ with $\alpha=0$. The final models are trained on these hyperparameters on different values of $\alpha$, in regular intervals from 1 to 8. The high values of $\alpha$ compared to \cite{rachidi} are justified by the ratio between $\loss$ and $\mathcal{R}$ for both regularization techniques.

All reported metrics, SHAP
analyses, figures, and tables are computed exclusively on
$\mathcal{D}^{(\rm test)}$, unless stated otherwise. This protocol
ensures a strict separation between model selection, auxiliary
supervision, and final evaluation.

\subsection{Baselines}

The selected features for the auxiliary logistic regression model are LDH and age. These are clinically significant
features because older patients tend to have poorer survival statistics for many cancer types \cite{age}, including MM \cite{riss}, and
LDH is one of the biomarkers used for R-ISS. 
The baselines include the auxiliary
logistic regression model, an ANN trained with the stage-based
regularizer only, and the unregularized ANN ($\alpha=0$). Table~\ref{table:baselines} reports accuracy and AUC for the three
baseline models on both $\mathcal{D}^{(\rm kf)}$ and the independent test
set $\mathcal{D}^{(\rm test)}$.

On $\mathcal{D}^{(\rm test)}$, the auxiliary model achieves the highest
performance, with accuracy $0.697$ and AUC $0.696$, closely followed by the
unregularized ANN. This indicates that a simple, clinically grounded
model based on age and LDH already provides strong predictive signal.
The stage-based baseline performs noticeably worse, suggesting that
coarse stage-level supervision alone is insufficient to capture the
full predictive structure of the data.

\begin{table}[h]
	\centering
	\caption{Metrics on baselines}
	\label{table:baselines}
	\begin{tabular}{|c|cc|cc|cc|}
		\hline
		& \multicolumn{2}{|c|}{\textbf{Auxiliary model}}  & \multicolumn{2}{|c|}{\textbf{Stage-based}}  & \multicolumn{2}{|c|}{$\bm{\alpha = 0}$} \\
		\hline
		\textbf{Dataset} & Acc & AUC & Acc & AUC & Acc & AUC\\
		\hline
		$\mathcal{D}^{(\rm kf)}$ & 0.646 & 0.645 & 0.634 & 0.627 & 0.736 & 0.734\\
		\hline
		$\mathcal{D}^{(\rm test)}$ & 0.697 & 0.696 & 0.631 & 0.621 & 0.689 & 0.686 \\
		\hline
	\end{tabular}
\end{table}

\subsection{Auxiliary-alignment regularization}

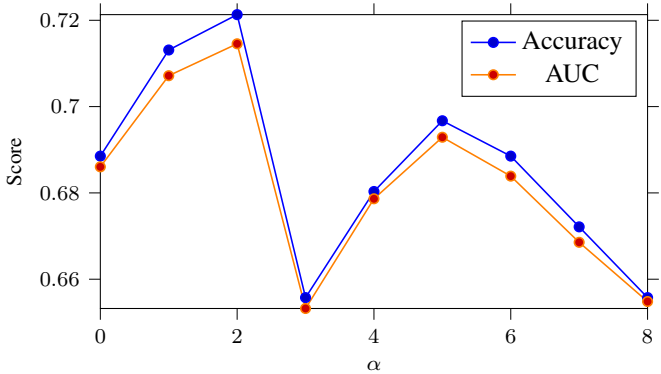
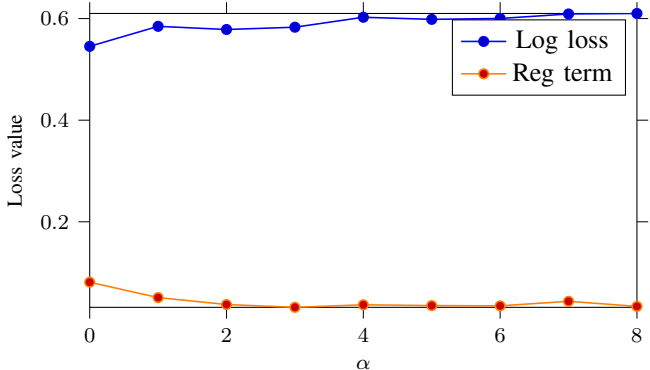
\begin{figure}[h]
	\centering
	\begin{subfigure}{\columnwidth}
		\centering
		\begin{tikzpicture}
			\begin{axis}[
				width=\columnwidth,
				height=0.62\columnwidth,
				xlabel={\footnotesize $\alpha$},
				ylabel={\footnotesize Score},
				axis line style={black},
				tick style={black},
				tick align=outside,
				xlabel near ticks,
				ylabel near ticks,
				xticklabel style={font=\footnotesize},
				yticklabel style={font=\footnotesize},
				label style={font=\footnotesize},
				every axis plot/.append style={black, semithick},
				enlargelimits=false,
				clip=true
				]
				\addplot+[
				mark=*,
				mark size=1.8pt,
				color=blue,
				semithick,
				error bars/.cd,
				y dir=both,
				y explicit,
				error bar style={semithick, blue},
				] table[
				x=alpha,
				y=accuracy,
				col sep=comma
				]{kfcv_lr/logreg_metr_test.csv};
				\addlegendentry{Accuracy}
				\addplot+[
				mark=*,
				mark size=1.8pt,
				color=orange,
				semithick,
				error bars/.cd,
				y dir=both,
				y explicit,
				error bar style={semithick, blue},
				] table[
				x=alpha,
				y=auc,
				col sep=comma
				]{kfcv_lr/logreg_metr_test.csv};
				\addlegendentry{AUC}
			\end{axis}
		\end{tikzpicture}
		\caption{Accuracy and AUC on $\mathcal{D}^{(\rm test)}$ for ANNs trained on different values of $\alpha$.}
		\label{fig:lr_metr_test}
	\end{subfigure}
	
	\begin{subfigure}{\columnwidth}
		\centering
		\begin{tikzpicture}
			\begin{axis}[
				width=\columnwidth,
				height=0.62\columnwidth,
				xlabel={\footnotesize $\alpha$},
				ylabel={\footnotesize Loss value},
				axis line style={black},
				tick style={black},
				tick align=outside,
				xlabel near ticks,
				ylabel near ticks,
				xticklabel style={font=\footnotesize},
				yticklabel style={font=\footnotesize},
				label style={font=\footnotesize},
				every axis plot/.append style={black, semithick},
				enlargelimits=false,
				clip=true
				]
				\addplot+[
				mark=*,
				mark size=1.8pt,
				color=blue,
				semithick,
				error bars/.cd,
				y dir=both,
				y explicit,
				error bar style={semithick, blue},
				] table[
				x=alpha,
				y=loss1,
				col sep=comma
				]{kfcv_lr/logreg_loss_test.csv};
				\addlegendentry{Log loss}
				\addplot+[
				mark=*,
				mark size=1.8pt,
				color=orange,
				semithick,
				error bars/.cd,
				y dir=both,
				y explicit,
				error bar style={semithick, blue},
				] table[
				x=alpha,
				y=reg_loss,
				col sep=comma
				]{kfcv_lr/logreg_loss_test.csv};
				\addlegendentry{Reg term}
			\end{axis}
		\end{tikzpicture}
		\caption{Dependency of logistic loss (blue) and regularizer $\mathcal{R}^{(\rm AA)}$ (orange), 
		evaluated on the test set $\mathcal{D}^{(\rm test)}$, on the regularization parameter $\alpha$.}
		\label{fig:logreg_loss_test}
	\end{subfigure}
	\caption{Comparison of scores and loss values with respect to $\alpha$ for models 
		using the auxiliary-alignment regularizer \eqref{eq:aux_reg}.}
	\label{fig:combined_panel}
\end{figure}

We first study the effect of the auxiliary-alignment regularizer
\eqref{eq:aux_reg}. Figure~\ref{fig:lr_metr_test} shows the accuracy and
AUC on $\mathcal{D}^{(\rm test)}$ as functions of $\alpha$. These predictive performance values are comparable to, and in
some cases slightly better than, the unregularized baseline. This
demonstrates that aligning the ANN with the auxiliary interpretable
model does not substantially deteriorate predictive performance. 

Figure~\ref{fig:logreg_loss_test} shows the logistic loss $\loss$ and 
regularizer value $\mathcal{R}^{(\rm AA)}$, evaluated on $\mathcal{D}^{(\rm test)}$, 
obtained for the ANN trained via \eqref{equ_def_rerm} for different values of $\alpha$.
As $\alpha$ increases, the regularization term $\mathcal{R}$ decreases 
monotonically, while the logistic loss $\loss$ increases gradually. The most pronounced change occurs between
$\alpha=0$ and $\alpha=1$, after which the trade-off stabilizes. This
behavior confirms that the regularizer effectively enforces auxiliary
alignment while preserving predictive performance over a wide range of
$\alpha$.

Interpretability effects are summarized in Table~\ref{table:shap_lr}, which 
reports SHAP-based feature rankings for different values of $\alpha$. Even for 
small $\alpha$, the regularizer shifts feature importance toward age and LDH, 
the two variables used in the auxiliary model $\widehat{g}$ 
and well-established clinical risk factors. In contrast, these features are ranked second and third in the unregularized ANN (using \eqref{equ_def_rerm} 
with $\alpha=0$).

\begin{table}[H]
	\centering
	\caption{Summary of SHAP value ranks for the auxiliary-alignment regularizer. The columns show which features were ranked 1st, 2nd or 3rd for each value of alpha according to the mean absolute Shapley values on $\mathcal{D}^{(\rm test)}$. 'lambda' and 'kappa' correspond to the respective free light chains.}
	\label{table:shap_lr}
	\begin{tabular}{|c|c|c|c|}
		\hline
		Rank & 1st & 2nd & 3rd \\
		$\alpha$ & & &\\
		\hline
		0 & lambda & age & LDH\\
		\hline
		1 & age & LDH & kappa\\
		\hline
		2 & age & LDH & albumin\\
		\hline
		3 &  age & LDH & hemoglobin\\
		\hline
		4 &  age & LDH & hemoglobin \\
		\hline
		5 &  age & LDH & kappa\\
		\hline
		6 &  age & LDH & hemoglobin \\
		\hline
		7 &  age & LDH & kappa\\
		\hline
		8 &  age & LDH & kappa\\
		\hline
	\end{tabular}
\end{table}

\subsection{Stage-consistency regularization}

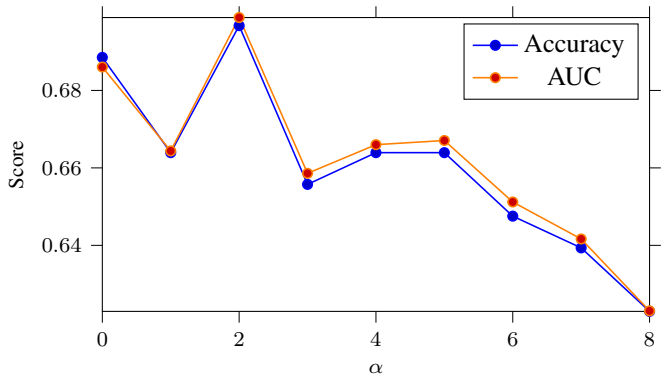
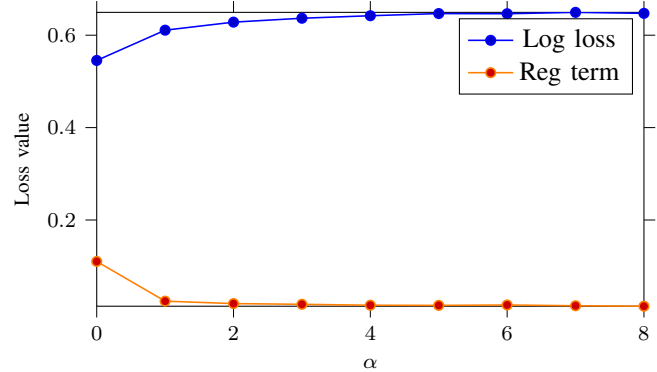
\begin{figure}[h]  
	\centering
	\begin{subfigure}{\columnwidth}
		\centering
		\begin{tikzpicture}
			\begin{axis}[
				width=\columnwidth,
				height=0.62\columnwidth,
				xlabel={\footnotesize $\alpha$},
				ylabel={\footnotesize Score},
				axis line style={black},
				tick style={black},
				tick align=outside,
				xlabel near ticks,
				ylabel near ticks,
				xticklabel style={font=\footnotesize},
				yticklabel style={font=\footnotesize},
				label style={font=\footnotesize},
				every axis plot/.append style={black, semithick},
				enlargelimits=false,
				clip=true
				]
				\addplot+[
				mark=*,
				mark size=1.8pt,
				color=blue,
				semithick,
				error bars/.cd,
				y dir=both,
				y explicit,
				error bar style={semithick, blue},
				] table[
				x=alpha,
				y=accuracy,
				col sep=comma
				]{kfcv_stage/stage_metr_test.csv};
				\addlegendentry{Accuracy}
				\addplot+[
				mark=*,
				mark size=1.8pt,
				color=orange,
				semithick,
				error bars/.cd,
				y dir=both,
				y explicit,
				error bar style={semithick, blue},
				] table[
				x=alpha,
				y=auc,
				col sep=comma
				]{kfcv_stage/stage_metr_test.csv};
				\addlegendentry{AUC}
			\end{axis}
		\end{tikzpicture}
		\caption{Accuracy and AUC on $\mathcal{D}^{(\rm test)}$ for ANNs trained on different values of $\alpha$.}
		\label{fig:stage_metr_test}
	\end{subfigure}
	
	\begin{subfigure}{\columnwidth}
		\centering
		\begin{tikzpicture}
			\begin{axis}[
				width=\columnwidth,
				height=0.62\columnwidth,
				xlabel={\footnotesize $\alpha$},
				ylabel={\footnotesize Loss value},
				axis line style={black},
				tick style={black},
				tick align=outside,
				xlabel near ticks,
				ylabel near ticks,
				xticklabel style={font=\footnotesize},
				yticklabel style={font=\footnotesize},
				label style={font=\footnotesize},
				every axis plot/.append style={black, semithick},
				enlargelimits=false,
				clip=true
				]
				\addplot+[
				mark=*,
				mark size=1.8pt,
				color=blue,
				semithick,
				error bars/.cd,
				y dir=both,
				y explicit,
				error bar style={semithick, blue},
				] table[
				x=alpha,
				y=loss1,
				col sep=comma
				]{kfcv_stage/stage_loss_test.csv};
				\addlegendentry{Log loss}
				\addplot+[
				mark=*,
				mark size=1.8pt,
				color=orange,
				semithick,
				error bars/.cd,
				y dir=both,
				y explicit,
				error bar style={semithick, blue},
				] table[
				x=alpha,
				y=reg_loss,
				col sep=comma
				]{kfcv_stage/stage_loss_test.csv};
				\addlegendentry{Reg term}
			\end{axis}
		\end{tikzpicture}
		\caption{Dependency of logistic loss (blue) and regularizer $\mathcal{R}^{(\rm SC)}$ (orange), 
			evaluated on the test set $\mathcal{D}^{(\rm test)}$, on the regularization parameter $\alpha$.}
		\label{fig:stage_loss_test}
	\end{subfigure}
	\caption{Comparison of scores and loss values with respect to $\alpha$ for models 
	        using the stage-consistency regularizer \eqref{equ:stage_reg}.}
	\label{fig:combined_stage_panel}
\end{figure}

We next examine the stage-consistency regularizer \eqref{equ:stage_reg}. 
Figure~\ref{fig:stage_metr_test} reports accuracy and AUC on $\mathcal{D}^{(\rm test)}$ 
as functions of $\alpha$ in \eqref{equ_def_rerm}. In
contrast to the auxiliary-alignment case, predictive performance
degrades steadily as $\alpha$ increases, with the exception of a small
performance recovery around $\alpha=2$. Overall, both accuracy and AUC
remain below the unregularized baseline for most values of $\alpha$.

Figure \ref{fig:stage_loss_test} depicts the logistic loss $\loss$ and 
regularizer value $\mathcal{R}^{(\rm SC)}$, evaluated on $\mathcal{D}^{(\rm test)}$, 
obtained for the ANN trained via \eqref{equ:stage_reg} for different values of $\alpha$.
As expected, increasing $\alpha$
reduces the regularization term $\mathcal{R}$ while increasing the
logistic loss $\loss$, with the largest change again occurring between
$\alpha=0$ and $\alpha=1$. The higher performance degradation compared
to Figure~\ref{fig:logreg_loss_test} indicates that enforcing 
within-stage consistency imposes a stronger constraint (via \eqref{equ:stage_reg}), 
compared to aligning with an auxiliary model (via \eqref{eq:aux_reg}). 

Table~\ref{table:shap_stage} summarizes SHAP-based feature rankings for
the stage-consistency regularizer. As $\alpha$ increases, the model
progressively emphasizes LDH, albumin, and B2M—features directly used
in R-ISS staging. This confirms that the regularizer successfully
injects the intended clinical structure. However, this emphasis comes
at the cost of reduced sensitivity to within-stage heterogeneity,
which likely explains the observed loss in predictive performance.

\begin{table}[h]
	\centering
	\caption{Summary of SHAP value ranks for the stage-consistency regularizer. The columns show the ranks of the features used in Algorithm \ref{alg_r_iss} for each value of alpha according to the mean absolute Shapley values on $\mathcal{D}^{(\rm test)}$.}
	\label{table:shap_stage}
	\begin{tabular}{|c|c|c|c|c|}
		\hline
		Feature & LDH & Albumin & B2M & Age \\
		$\alpha$ & & & & \\
		\hline
		0 & 3rd  & 5th & 8th & 2nd \\
		\hline
		1 & 1st & 2nd & 5th & 3rd\\
		\hline
		2 & 1st & 2nd & 3rd & 7th \\
		\hline
		3 &  1st & 2nd & 3rd & 7th \\
		\hline
		4 &  1st & 2nd &  3rd & 11th \\
		\hline
		5 &  1st & 2nd & 3rd & 11th\\
		\hline
		6 & 1st & 2nd & 4th & 12th \\
		\hline
		7 & 1st & 2nd & 3rd & 11th \\
		\hline
		8 & 1st & 2nd & 3rd & 9th\\
		\hline
	\end{tabular}
\end{table}

\section{Conclusion} \label{conclusion}

We studied two interpretability-driven regularization strategies for five-year
survival prediction in patients with MM. The results show that
interpretability constraints can be incorporated directly into model
training without substantially degrading predictive performance, when
the imposed structure aligns well with clinically meaningful notions of
similarity. In particular, auxiliary-alignment regularization provides 
a favorable trade-off between predictive accuracy and interpretability, 
whereas coarser stage-based consistency constraints reduce model flexibility
and predictive performance. These findings highlight the importance of
carefully selecting clinically grounded structures when designing
interpretability-aware learning objectives.







\bibliographystyle{IEEEtran}

\end{document}